\documentclass{article}

\PassOptionsToPackage{numbers, compress}{natbib}
\usepackage[preprint]{neurips_2026}

\usepackage[utf8]{inputenc} 
\usepackage[T1]{fontenc}    
\usepackage[hidelinks]{hyperref} 
\usepackage{url}            
\usepackage{booktabs}       
\usepackage{amsfonts}       
\usepackage{nicefrac}       
\usepackage{microtype}      
\usepackage{xcolor}         
\usepackage{graphicx}
\usepackage{amsmath}
\usepackage{cleveref}
\usepackage{multirow}
\usepackage{subcaption}
\usepackage{algorithm}
\usepackage{algorithmic}
\usepackage{placeins} 

\title{COLLAR: Cascaded Object-Level Latent Refinement for High-Fidelity Conditional Generation}

\author{%
  Xinlong Zhang, Jia Wei, Xiaoyu Zhang, Teng Zhou, Chengyu Lin, Yongchuan Tang\thanks{Corresponding author.} \\
  College of Computer Science, Zhejiang University, Hangzhou, Zhejiang, China \\
  {\small\texttt{\{xinlzhang, weijia\_77, xiaoyzhang, tengzhou, linchengyu, yctang\}@zju.edu.cn}}
}

\begin{document}

\maketitle

\begin{abstract}
Achieving high-fidelity object-level control in Diffusion Transformers remains a significant challenge despite the introduction of structural priors like depth and Canny maps. Current object-level conditional generation methods frequently suffer from visual artifacts and struggle to maintain precise control over objects within small localized regions. To address these limitations, we propose Cascaded Object-Level Latent Refinement (COLLAR), a training-free framework that progressively optimizes object-level features via the Field-of-View (FoV) expansion. First, we propose the Cross-Scale Semantic Alignment (CSSA) module to address spatial-semantic gaps by injecting object-level features into extended-FoV branches via attention mechanisms. To further optimize these features, the Cyclic Feature Injection (CFI) module introduces a reciprocal background feedback mechanism. It leverages a frequency-based adaptive strategy to selectively update the global backbone with context-aligned local information. Finally, the extended-FoV branch serves as a hub for integrating context-aligned object-level features into the global generation process. Experiments on COCO-MIG and COCO-POS demonstrate competitive performance across spatial fidelity, semantic alignment, and image quality, with COLLAR achieving the strongest spatial alignment under Canny conditioning and the lowest  FID on COCO-MIG under depth conditioning.
\end{abstract}

\section{Introduction}
\label{sec:intro}

\begin{figure}[tb]
  \centering
  \includegraphics[height=6.5cm]{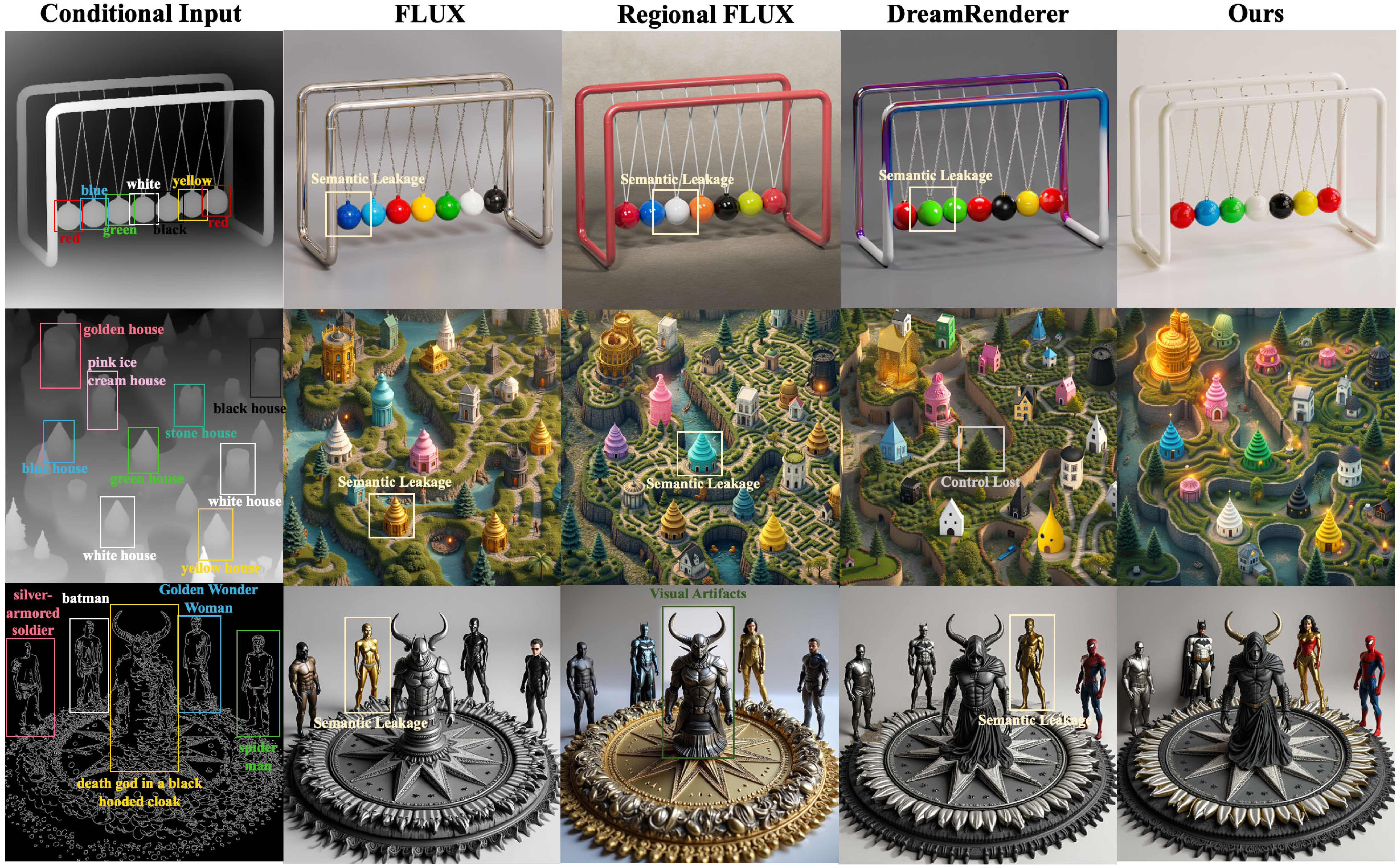}
  \caption{Our proposed framework, COLLAR, is a plug-and-play module specifically designed for object-level conditional generation. It facilitates precise, fine-grained semantic control across diverse spatial modalities. Notably, our approach excels at mitigating attribute leakage and enhancing the generation of small-scale objects.
  }
  \label{teaser}
\end{figure}

In recent years, text-to-image diffusion models \cite{c1, c2, c3} have achieved remarkable success in image generation. Building upon their generative capability, the integration of spatial cues further empowers models to grasp underlying scene structures \cite{c4, c5, c6}, thereby enabling precise pixel-level control. Such controllability is indispensable for a wide range of practical applications, including professional design\cite{c7, c8}, advertising\cite{c11}, and animation\cite{c9, c10}. However, as the demand for fine-grained control grows, relying solely on text prompts and a single conditional image often falls short in complex, multi-object scenarios. As illustrated in \Cref{teaser}, existing methods still struggle with unavoidable semantic leakage. Specifically, when the global structure contains similar objects, the model fails to accurately render specified attributes (e.g., color) onto the target regions.

To alleviate the above problems, existing studies \cite{c12, c13, c14, c15} typically introduce layout information to achieve object-level control in complex scenarios. These methods are mainly divided into two categories: training-free \cite{c15, c16, c17} and training-based methods \cite{c13, c14, c18}. Training-free methods include pixel-level feature aggregation and attention-based feature modulation. Pixel-level feature aggregation \cite{c55, c56} offers strong control over object locations. Due to distribution discrepancies across features of distinct objects, the fusion process is prone to distribution bias, which inevitably introduces artifacts into the final images \cite{c15, c16}. Attention-based manipulation \cite{c12, c17, c19} enables control by reinforcing semantic activations in targeted regions, but it exhibits instability when handling small-scale objects. Tuning-based methods \cite{c42, c43} use bounding-box-guided adapters to control object position and appearance. However, they introduce additional parameters, increasing training complexity and resource overhead. Moreover, such adapters struggle to generalize to other models. The limitations of prior work, as discussed above, motivate the need for further research to address these challenges in object-level conditional generation. 

In this paper, we propose a novel framework named COLLAR (\textbf{C}ascaded \textbf{O}bject-\textbf{L}evel \textbf{LA}tent \textbf{R}efinement). It is a training-free method to achieve high-fidelity object-level conditional image generation. Inspired by \cite{c50}, COLLAR introduces a paradigm of progressive feature refinement and adaptive fusion. Specifically, it consists of two core modules: (a) Cross-Scale Semantic Alignment (CSSA) module and (b) Cyclic Feature Injection (CFI) module. The CSSA module uses attention-based manipulation to dynamically integrate bounding-box-restricted local features into corresponding extended Field-of-View (FoV)  branches. This design strictly preserves local semantic alignment while enhancing the spatial contextual awareness of features. The CFI module establishes a reciprocal background feedback mechanism. It extracts global background features to extended-FoV branches and feeds object-level features into the global branch. Concurrently, we design a frequency-based adaptive injection strategy to selectively inject context-aligned object-level features into the global backbone. Serving as a central hub for feature optimization, extended-FoV branches allow object-level features to smoothly blend into the global generation process while strictly retaining core semantics. We extensively evaluate COLLAR on the COCO-POS and COCO-MIG benchmarks. Extensive quantitative and qualitative experiments demonstrate that COLLAR outperforms existing baselines across multiple metrics. It helps reduce visual artifacts while maintaining competitive image quality and spatial fidelity.

Our contributions are summarized as follows:
\begin{itemize}
\item We propose COLLAR, a training-free framework that achieves high-fidelity object-level conditional generation under global structural constraints.
\item We introduce the Cyclic Feature Injection technique that facilitates reciprocal feature feedback, enabling local object semantics to seamlessly blend into the global generation process.
\item We design an object-level adaptive injection strategy. By dynamically terminating feature injection, it helps reduce structural artifacts and improves inference speed.
\item Extensive quantitative and qualitative evaluations demonstrate that our approach outperforms existing methods in object-level conditional generation.
\end{itemize}

\section{Related Work}
\subsection{Text-to-Image Generation}
Text-to-image generation aims to learn a visual-semantic mapping from semantic space to visual representations \cite{c20, c21}. Recently, the probabilistic diffusion paradigm \cite{c22} has made significant progress in both generation fidelity and text-image alignment \cite{c23}. Building on this paradigm, Stable Diffusion \cite{c1} demonstrates significant improvements in generation efficiency and fidelity by denoising in a compressed latent space \cite{c1, c2}. SDXL \cite{c2} also follows this design philosophy. With the advent of the Diffusion Transformer (DiT) architecture \cite{c26, c27}, models such as SD3 \cite{c3} adopt transformer-based backbones with joint attention mechanisms to implicitly model cross-modal semantic correspondences. This design facilitates bidirectional interaction across different modalities, alleviating the information bottleneck inherent \cite{c4}. To handle complex text prompts, these models commonly use larger text encoders, such as T5 \cite{c28, c29}.

\subsection{Conditional Text-to-Image Generation}
To enhance T2I models' ability for spatial geometric structures, researchers have extended diffusion models to accommodate diverse spatial conditions \cite{c4, c5, c52}. Early studies were predominantly built upon the Stable Diffusion framework, with representative works including ControlNet \cite{c4} and T2I-Adapter \cite{c5}. These methods introduce trainable structural control modules that enable precise injection of geometric priors (e.g., depth maps \cite{c30}, Canny edges \cite{c38}, and segmentation maps \cite{c31}), while preserving the original generative capabilities of the pre-trained model. Subsequently, Uni-ControlNet \cite{c32} and UniControl \cite{c33} proposed a novel unified control paradigm by constructing universal conditional encoders capable of hybrid representations for multimodal control signals. To achieve multi-condition controllable generation, Cocktail \cite{c34} and Cross-ControlNet \cite{c35} have explored novel feature-aggregation strategies. As diffusion model architectures evolve toward DiT, approaches like FLUX-Depth \cite{c58} and EasyControl \cite{c6} leverage LoRA's \cite{c37} parameter efficiency to guide large-scale DiT networks in effectively learning complex spatial-structure priors while maintaining high image fidelity.

\subsection{Layout-to-Image Generation}
Layout-to-image (L2I) generation aims to achieve object-level compositional image synthesis conditioned on instance information \cite{c39}. Existing methods can be broadly categorized into training-based \cite{c14, c18, c40, c41, c42, c43}  and training-free approaches\cite{c15, c16, c19, c55, c56}. Training-based methods typically introduce novel adapters to model layout information for object-level controllable generation, such as GLIGEN \cite{c40}, InstanceDiffusion \cite{c41}, and CreatiLayout \cite{c14}. Training-free methods often explicitly manipulate cross-attention or joint attention layers to associate object descriptions with local image features, thereby enabling multi-object control. For example, DreamRenderer introduced a Hard Text Attribute Binding mechanism to ensure that the text embedding of each instance correctly aligns with its visual information \cite{c12}.   Despite their effectiveness in controlling fine-grained attributes such as object position, color, and style, some of these methods remain prone to semantic leakage and edge artifacts when combined with global spatial conditioning.

\section{Methods}
\subsection{Preliminaries}
FLUX is a state-of-the-art diffusion-based text-to-image generation model. Its core architecture leverages the T5 text encoder to map complex textual prompts into high-dimensional sequence embeddings. These embeddings are subsequently fused with image features via multimodal joint-attention layers, enabling bidirectional interaction across modalities. FLUX is built upon the multimodal DiT, enabling the semantic alignment of image and text embeddings within a unified embedding space. During the generation process, image features are iteratively denoised via Flow Matching and then decoded into high-fidelity images by a pre-trained autoencoder. Furthermore, FLUX supports conditional image generation tasks—such as FLUX-Depth and FLUX-Canny—by concatenating control images (encoded by the VAE) with the latent features along the channel dimension, thereby achieving precise spatial guidance.

In each MM-DiT layer, image and text features are first normalized via adaptive Layer Normalization (AdaLN), where the normalization parameters are derived from fused timestep and CLIP text embeddings. The normalized image and text features are projected into queries, keys, and values. The corresponding features are then concatenated along the sequence dimension. Before being fed into the self-attention module, the query and key features are augmented with rotary position embeddings (RoPE). This process can be formalized as follows:

\begin{align}
  Q= \text{RoPE}(\left [ Q^{T}, Q^{I} \right ]),  \qquad   &K= \text{RoPE}(\left [ K^{T}, K^{I} \right ]) ,  \qquad   V= \left [ V^{T}, V^{I} \right ],\\
  \text{Attention}& \left (Q, K, V \right ) = \text{Softmax}(\frac{QK^{T} }{\sqrt{d} })V.
  \label{eq2}
\end{align}

\noindent where $Q^{I}, K^{I}, V^{I}$ denote the query, key, and value features for the image modality, and $Q^{T}, K^{T}, V^{T}$ for the text modality. $d$ is the dimension of the key vectors.
This mechanism facilitates bidirectional interaction between modalities, enabling better alignment between text and image features.

\begin{figure}[tb]
  \centering
  \includegraphics[height=6.5cm]{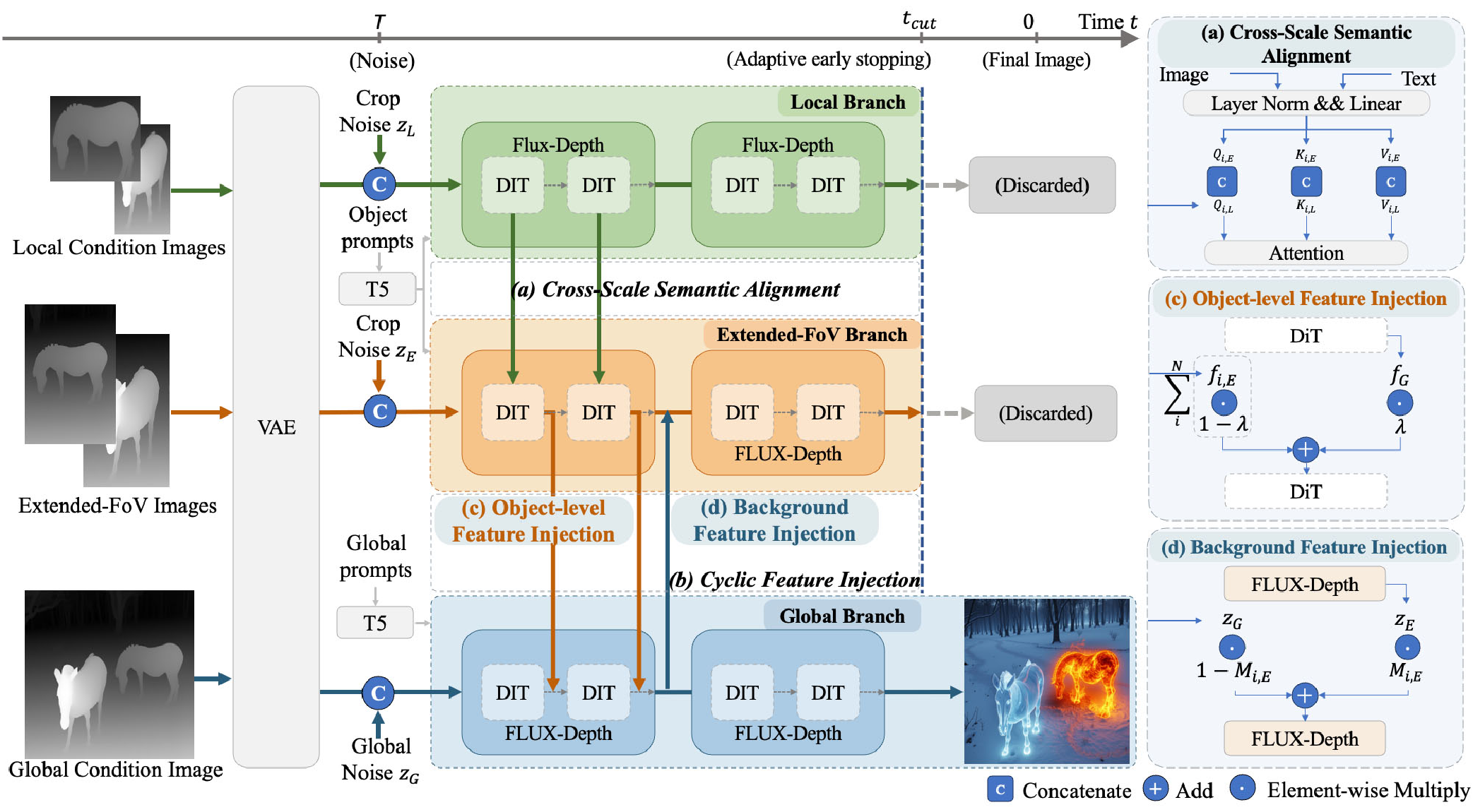}
  \caption{Overview of COLLAR. We freeze the FLUX-Depth and use COLLAR to achieve object-level conditional T2I generation. COLLAR contains two processing modules. By progressively optimizing object-level features, our method adaptively integrates them into the global generation process while helping maintain image quality.
  }
  \label{fig2}
\end{figure}

\subsection{Pipeline}
As illustrated in \Cref{fig2}, we propose a progressive feature optimization and fusion strategy to improve local fidelity across different instances. Our pipeline consists of two core modules: (i) a Cross-Scale Semantic Alignment (CSSA) module, and (ii) a Cyclic Feature Injection (CFI) module. In the CSSA module (\Cref{fig2}(a)), we use semantic features from local branches to guide the generation of extended-field-of-view branches, making it take into account both details and context.  Specifically, the conditional image crops within each bounding box serve as inputs for local branches to generate semantically aligned details.  The extended-FoV branch leverages the fine-grained features from local branches as a semantic reference to dynamically reinforce its own object-level representations during denoising. In the CFI module (\Cref{fig2}(b)), we explicitly align local features with the global background, enabling the seamless integration of instances without compromising fidelity.   Specifically, we extract background features from the global branch that spatially align with the extended region of the extended-FoV branch.   Subsequently, these extracted features are injected into the extended-FoV branch to smooth object semantics. Finally, we design an object-level early-stopping strategy, and the global branch adaptively integrates these object features to synthesize the final high-fidelity image. Serving as a central harmonization hub, the extended branch effectively bridges the spatial scales, and the resulting harmonized object representations are finally fed back into the global backbone.

\subsection{Cross-Scale Semantic Alignment}
\label{sec34}
Given the impressive performance of pretrained models, existing object-level generation methods typically extract instance-conditioned features and then fuse them into the global stream using bounding boxes. However, these methods exhibit inherent limitations when global spatial conditions are introduced to guide the generation process. For small-scale objects, the inherently limited resolution of local conditional inputs is further exacerbated by the VAE encoder's downsampling, leading to a severe loss of fine-grained spatial details. As a result, generated instances often lack fine details and fail to achieve precise semantic alignment with their textual descriptions, thereby compromising subsequent multi-instance feature fusion. Expanding the field of view provides more spatial context for object generation. However, background content and nearby objects may weaken alignment with the instance prompt.

Therefore, we propose the CSSA module, which explicitly injects object features from the local branch into the extended-FoV branch to enhance semantics.
To achieve this, we employ an attention-based feature fusion mechanism, as illustrated in \Cref{fig2}(a). We extract the $Q_{i, L}$, $K_{i, L}$, and $V_{i, L}$ from the local branch corresponding to the $i$-th instance, and feed them into the DiT layer of the corresponding extended-FoV branch. Within this layer, these features are concatenated with extended-branch features $Q_{i, E}$, $K_{i, E}$, and $V_{i, E}$ along the sequence dimension as follows:

\begin{align}
	&\widetilde{Q}_{i,E}= [\text{RoPE}_{E}(Q_{i,E}), \text{RoPE}_{L}(Q_{i,L})], \\
	&\widetilde{K}_{i,E}= [\text{RoPE}_{E}(K_{i,E}), \text{RoPE}_{L}(K_{i,L})], \\
	&\widetilde{V}_{i,E}= \left [ V_{i,E}, V_{i,L} \right ], \\
	&\text{RoPE}_{L} = \text{Crop}(\text{RoPE}_{E}, b_{i})
\end{align}

\noindent $\text{RoPE}_{E}$ and $\text{RoPE}_{L}$ respectively represent rotary position embeddings for the extended-FoV branch and the local branch, and $b_{i}$ represent the bounding box of the instance $i$. Then, these concatenated features are fed into \Cref{eq2} to execute feature fusion. It is worth noting that in the DiT layer, text and image features interact densely, and text features also encode visual information. To fully preserve the semantic information from local branches, we feed both their text and image features into the extended-FoV branch. Crucially, we explicitly align the position embeddings by treating the extended-FoV branch as the reference coordinate space. Specifically, position embeddings for features from the local branch are directly assigned the exact spatial subset of indices corresponding to their shared overlapping region. This strict relative alignment empowers the extended branch to precisely localize and attend to the local priors.  Furthermore, we also support initializing local branches with distinct noise latents for object $i$.  By aggregating features from multiple parallel local branches and injecting them into the extended-FoV branch, the semantic information is  enriched.

\subsection{Cyclic Feature Injection}
Considering that the extension-FoV branch is still guided by the instance’s prompt, forcefully integrating optimized local features (obtained by \Cref{sec34}) directly into global features can easily lead to visual disharmony.  Therefore, we propose the CFI module.  To clarify this module, we formalize it step by step.

\noindent \textbf{Background Feature Injection}.  First, we extract the background feature from the global branch that spatially aligns with the extended region of the extended-FoV branch (\Cref{fig2}(d)).  Let ${b}_{i}$ denote the initial bounding box of the object $i$, and we can obtain the extended-FoV bounding box, ${b}_{i, E}$, by applying a spatial expansion operation with a scale factor $\gamma$:
 
\begin{align}
{b}_{i, E} = \text{Expand}(b_{i}, \gamma)
\end{align}

\noindent Then, we crop the corresponding regional features from the global branch $z_{G}$:

\begin{align}
z_{E}^{global} = \text{Crop}(z_{G}, b_{i, E})
\end{align}

\noindent Finally, we replace the background features in the extended-FoV branch with the cropped global features while retaining the features within the object region:
\begin{align}
z_{E} \leftarrow z_{E}^{global} \odot (1 - M_{i, E}) + z_{E} \odot M_{i, E}
\end{align}
\noindent where $M_{i, E}$ represents the object region in the extended-FoV branch.

\noindent \textbf{Object-level Feature Injection}. Following refinement in the CSSA module and background-feature injection (\Cref{fig2}(d)), object-level features in the extended-FoV branch seamlessly fuse local semantics with spatial context. Consequently, we extract these features and precisely inject them into the corresponding regions of the global branch (\Cref{fig2}(c)). Specifically, at the output of each selected DiT layer, the global feature $f_{G}$ is updated via a weighted summation with the feature set $\{f_{i, E}\}_{1}^{N}$ extracted from extended-FoV branches:
\begin{align}
f_{G} \leftarrow f_{G}\odot(1-M_G) + \left[\lambda f_{G} + \frac{1-\lambda}{\max\!\left(\sum_{i=1}^{N}M_{i,G},1\right)}\sum_{i=1}^{N}\left(f_{i,E}\odot M_{i,G}\right)\right]\odot M_G
\label{eq:ofi}
\end{align}
\noindent where $M_{i,G}$ denotes the globally aligned mask of object $i$, and $M_G$ is the union of all such masks. The weighted fusion is applied only within the object regions, while features outside these regions are entirely provided by the global branch. Due to the spatial resolution mismatch between the extended-FoV branch and the global branch, we explicitly zero-pad the extracted features $f_{i, E}$ and map their masks to the global feature grid before performing weighted summation.

\noindent \textbf{Object-level Early-Stopping Strategy}. The guidance step of object-level feature injection significantly impacts the final image quality. Insufficient steps lead to inadequate semantic activation, whereas excessive injection compromises global coherence. To achieve a trade-off, we propose an object-level early-stopping strategy. At each timestep $t$, we employ the masks $M_{i, G}$ and $M_{i, E}$ to extract the local latent features from the global and the extended-FoV branch $i$, respectively. These features are then processed by a high-pass filter $\mathcal{H}(\cdot)$:

\begin{align}
H_{i, G}^{(t)} = \mathcal{H}(z_{G}^{(t)} \odot M_{i, G}), \quad H_{i, E}^{(t)} = \mathcal{H}(z_{i, E}^{(t)} \odot M_{i, E})
\end{align}

We compute token-wise cosine similarities between the resulting high-frequency components:
\begin{align}
{s}_i^{(t)} = \text{CosSim}(H_{i, G}^{(t)}, H_{i, E}^{(t)})
\end{align}

\noindent The global and extended-FoV branches share initial latents in corresponding object regions. Their similarities can therefore be high early in the denoising process. We apply min--max normalization to the token-wise similarities within each object region. Lower normalized values indicate positions with relatively lower similarity between the branches. We average these values to obtain the final stopping score $s_i^{(t)}$. The predefined threshold $\sigma$ is set to 0.85 by default. If $s_i^{(t)} \ge \sigma$, the feature injection for instance $i$ is terminated at the timestep $t$ in the global branch (\Cref{fig2}). The intuition behind this design is that high-pass residuals are sensitive to local edges and contours, as supported by the analyses in \Cref{sec:highpass-validation}. As denoising progresses, continued object-level feature injection may disrupt global coherence and introduce visual artifacts. Overall, this strategy significantly reduces inference costs while helping the final images remain natural and high-quality. In addition, we set the maximum guidance step $S$.

\section{Experiments}
\label{sec:blind}
\subsection{Implementation Details}
\label{Impdetails}
To evaluate the effectiveness of COLLAR, we conduct experiments on FLUX.1-Depth and FLUX.1-Canny \cite{c58}.  During inference, the FlowMatchEulerDiscreteScheduler is employed with 20 sampling steps for all benchmarks.  For structural guidance, the Classifier-Free Guidance scale is set to 10.0 for depth-conditioned generation and 30.0 for Canny-conditioned generation.  All experiments are conditioned on boxes, object-level descriptions, text prompts, and structural images.

\subsection{BaseLines}
We compare COLLAR with state-of-the-art (SOTA) training-free, object-level conditional generation methods, including DreamRenderer \cite{c12}, 3DIS \cite{c17}, and Regional-FLUX \cite{c15}.  Given that COLLAR employs a latent-level patch transplantation strategy, we also include several representative methods, such as NoiseCollage \cite{c55} and GrounDit \cite{c56}. To ensure a fair comparison, we omit the plug-and-play strategy R\&B \cite{c59} from GrounDit, as it is used for the cross-attention module and is not directly applicable to FLUX's joint-attention architecture. Furthermore, we also integrate COLLAR into the latest conditional generation models, such as EasyControl \cite{c6} and SD3 \cite{c3}, to demonstrate its compatibility and effectiveness.

\subsection{Evaluation Benchmarks}
We evaluate the effectiveness of COLLAR using the COCO-POS and COCO-MIG \cite{c12}. We randomly select 300 samples from the COCO-2014 dataset \cite{c51}.   For each sample, 8 random seeds are used for generation, yielding a total of 2,400 images per method for evaluation. In COCO-MIG and COCO-POS, we assess local fidelity with respect to attribute binding, spatial accuracy, semantic consistency, and image quality under depth and Canny maps. In \Cref{t1} and \Cref{t2}, we examine COLLAR's integration versatility. Specifically, we conduct a comparative analysis between the original backbone models and their COLLAR-enhanced versions.

\subsection{Evaluation Metrics}
We use the following metrics for measuring the performance of different models: (1) \textbf{Spatial Fidelity}: mean Intersection over Union (mIoU) and mean Average Precision (mAP) are employed to assess the grounding accuracy between synthesized objects and the prescribed layouts. To extract object-level bounding boxes for evaluation, we utilize SAM3 \cite{c53} as an off-the-shelf detector. Furthermore, the Object Success Ratio (OSR) and the Image Success Ratio (ISR) are computed to quantitatively evaluate multi-instance generation. OSR denotes the ratio of correctly rendered instances, while ISR indicates the percentage of images where all objects are correctly rendered. For COCO-POS (\Cref{tab1}), ISR and OSR evaluate category correctness only, without considering color attributes. For COCO-MIG (\Cref{tab2}), a successful instance must have both the correct category and color attribute; (2) \textbf{Semantic Consistency}: Beyond spatial accuracy, we employ Global CLIP (G-CLIP), Local CLIP (L-CLIP), HPSv3 \cite{c54}, and PickScore \cite{c57} to evaluate text-image alignment; (3) \textbf{Image Quality}: We utilize the Fréchet Inception Distance (FID) to evaluate the visual quality and photorealism of the synthesized images.

\subsection{Quantitative Comparison}
\Cref{tab1} and \Cref{tab2} collectively present a comprehensive evaluation of our proposed COLLAR, demonstrating its effectiveness in balancing precise spatial condition adherence with high generative fidelity. As detailed in \Cref{tab1}, under the Depth modality, our method achieves the best FID (16.3) and highly competitive aesthetic performance, ranking second in HPSv3 (6.15). On COCO-MIG under depth conditioning, NoiseCollage achieves slightly higher OSR and mIoU than COLLAR, but has a higher FID of 20.4 compared with 16.3 for COLLAR. Its advantage in spatial alignment may come from the dense spatial cues in depth maps. The robustness of COLLAR is more evidently observed under the challenging Canny modality. Without dense spatial cues, several baseline models struggle to maintain structural integrity. In contrast, COLLAR achieves the highest OSR (68.9) and ISR (40.8) on COCO-POS. On COCO-MIG, it obtains the highest PickScore (0.153) and a competitive FID of 18.2. To further validate this capability, \Cref{tab2} provides a fine-grained evaluation on multi-instance generation. As the scene complexity increases from $n_2$ to $n_6$ objects, methods such as Regional-FLUX and GrounDit experience noticeable performance degradation. In contrast, COLLAR generally maintains the highest or second-highest success rates across both modalities. Particularly under the Canny constraint, our approach achieves the highest OSR across all instance tiers while \Cref{tab1} reports the highest overall COCO-MIG OSR (51.8) and ISR (22.4).

\begin{table}[tb]
  \caption{Quantitative comparisons on the COCO-POS and COCO-MIG benchmarks. COCO-POS spatial metrics evaluate category correctness only, whereas COCO-MIG spatial metrics require both the correct category and attribute. \textbf{Bold} and \underline{underline} represent the best and second-best methods, respectively. The two Canny FIDs reported as 18.2 are not assigned a second-place marker.}
  \label{tab1}
  \scriptsize
  \centering
  \setlength{\tabcolsep}{2.5pt}
  \begin{tabular}{@{}llccccccccccccc@{}}
    \toprule
    & & \multicolumn{4}{c}{COCO-POS} & \multicolumn{9}{c}{COCO-MIG} \\
    \cmidrule(lr){3-6} \cmidrule(lr){7-15}
    Input & Method & ISR$\uparrow$ & OSR$\uparrow$ & mIoU$\uparrow$ & mAP$\uparrow$ & G-CLIP$\uparrow$ & L-CLIP$\uparrow$ & ISR$\uparrow$ & OSR$\uparrow$ & mIoU$\uparrow$ & mAP$\uparrow$ & FID$\downarrow$ & HPSv3$\uparrow$ & Pick$\uparrow$ \\
    \midrule
    \multirow{7}{*}{\rotatebox[origin=c]{90}{\bf Depth}} & FLUX & 62.6 & 81.7 & 74.9 & 83.2 & 31.40 & 26.97 & 24.1 & 46.0 & 43.7 & 47.5 & / & 6.14 & \textbf{0.156} \\
                                     & 3DIS & 63.4 & 82.9 & 75.5 & 83.8 & 31.46 & 27.85 & 42.3 & 65.9 & 61.0 & 68.2 & 18.3 & 5.88 & 0.143 \\
                                     & Regional-FLUX & 64.9 & 83.4 & 76.0 & 84.9 & 30.79 & 27.65 & 37.6 & 63.8 & 59.2 & 66.0 & 19.8 & 6.00 & 0.135 \\
                                     & NoiseCollage & 65.7 & 84.0 & 76.4 & 85.9 & 31.58 & \textbf{28.23} & \textbf{47.8} & \textbf{71.7} & \textbf{66.0} & \textbf{74.9} & 20.4 & 5.83 & 0.140 \\
                                     & GrounDit & 65.3 & 83.9 & 75.8 & 85.3 & 31.21 & 27.98 & 43.0 & 69.3 & 63.3 & 71.1 & 19.6 & 5.64 & 0.126 \\
                                     & DreamRenderer & \underline{66.5} & \textbf{85.8} & \textbf{78.2} & \underline{86.6} & \textbf{31.73} & 28.15 & 42.0 & 68.1 & 63.1 & 70.0 & \underline{17.6} & \textbf{6.18} & \underline{0.151} \\
                                     & Ours & \textbf{66.7} & \underline{84.8} & \underline{76.9} & \textbf{86.7} & \underline{31.63} & \underline{28.16} & \underline{46.8} & \underline{71.0} & \underline{65.0} & \underline{73.8} & \textbf{16.3} & \underline{6.15} & 0.146 \\
    \midrule
    \multirow{7}{*}{\rotatebox[origin=c]{90}{\bf Canny}} & FLUX & 32.6 & 60.0 & 56.3 & 62.1 & 30.23 & 25.59 & 9.0 & 29.6 & 31.8 & 32.6 & / & 6.04 & 0.149 \\
                                     & 3DIS & 34.3 & 62.3 & 57.4 & 65.4 & 30.21 & 26.41 & 16.5 & 42.5 & 41.8 & 45.5 & 18.2 & 6.02 & 0.144 \\
                                     & Regional-FLUX & 36.4 & 64.5 & 59.3 & 66.5 & 29.63 & 26.41 & 15.5 & 42.8 & 41.9 & 45.2 & 19.3 & 5.91 & 0.132 \\
                                     & NoiseCollage & 34.6 & 63.0 & 57.8 & 65.1 & 30.36 & 26.43 & 16.2 & 43.7 & 42.3 & 46.8 & 20.7 & 5.68 & 0.139 \\
                                     & GrounDit & 36.8 & 64.9 & 59.2 & 67.3 & 30.27 & 26.81 & \underline{19.3} & \underline{47.1} & \underline{45.0} & \underline{49.6} & 19.2 & 5.86 & 0.131 \\
                                     & DreamRenderer & \underline{39.9} & \underline{67.3} & \underline{61.7} & \underline{69.4} & \underline{30.66} & \underline{26.93} & 17.5 & 45.4 & 44.3 & 47.8 & \textbf{17.3} & \textbf{6.25} & \underline{0.151} \\
                                     & Ours & \textbf{40.8} & \textbf{68.9} & \textbf{62.2} & \textbf{71.9} & \textbf{30.80} & \textbf{27.21} & \textbf{22.4} & \textbf{51.8} & \textbf{48.6} & \textbf{55.5} & 18.2 & \underline{6.24} & \textbf{0.153} \\
    \bottomrule
  \end{tabular}
\end{table}

 \begin{figure}[t]
  \centering
   \includegraphics[height=8cm]{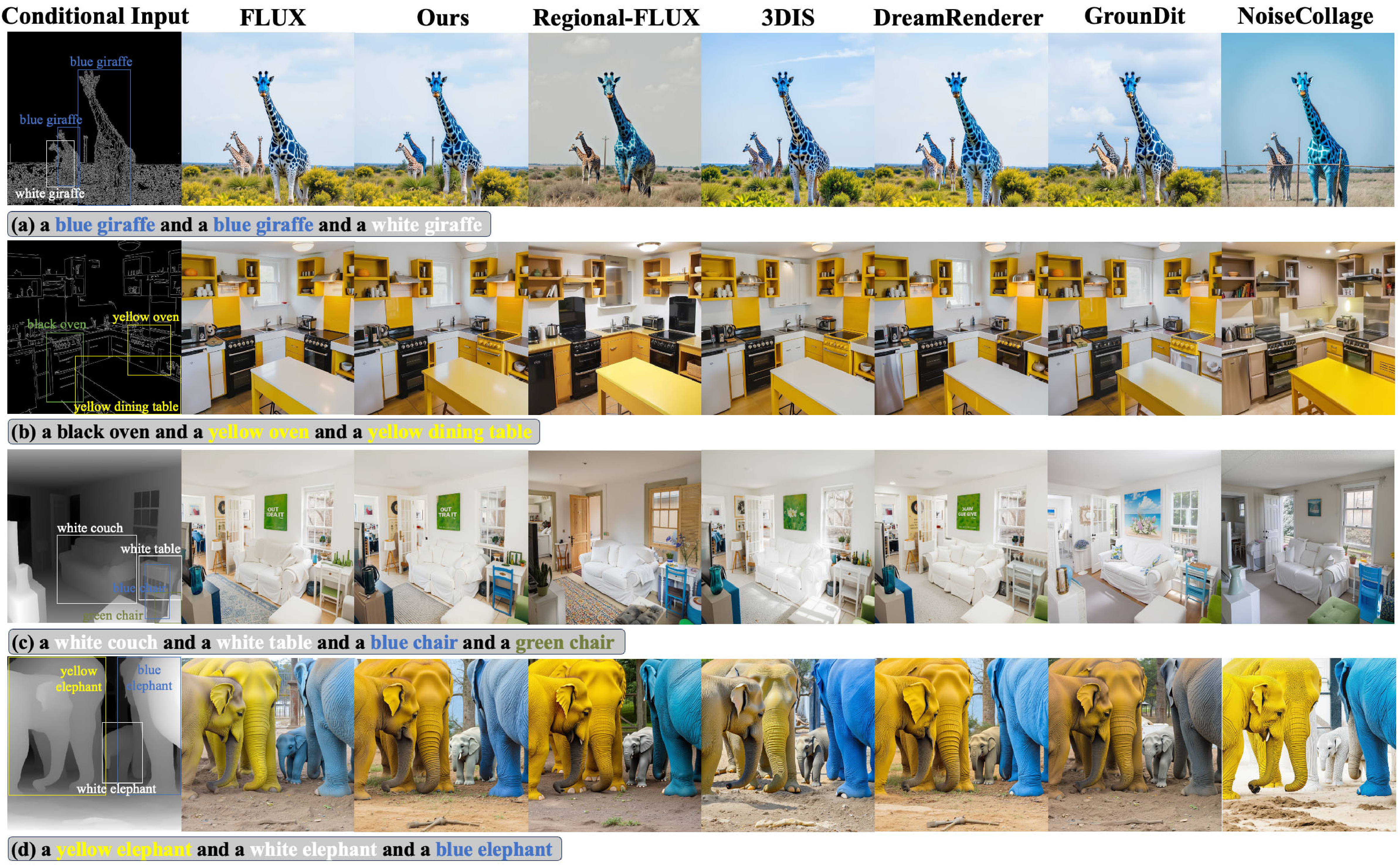}
   \caption{Qualitative comparison between our method and state-of-the-art baselines. Visualization results demonstrate that COLLAR effectively preserves local structural integrity while maintaining precise attribute alignment.}
   \label{fig3}
\end{figure}

\begin{table}[tb]
  \caption{Quantitative comparisons on the COCO-MIG benchmark.}
  \label{tab2}
   \scriptsize
  \centering
  \begin{tabular}{@{}llcccccccccc@{}}
    \toprule
     & & \multicolumn{5}{c}{Image Success Ratio$\uparrow$} & \multicolumn{5}{c}{Object Success Ratio$\uparrow$} \\
     \cmidrule(lr){3-7} \cmidrule(lr){8-12}
    Input & Method & \multicolumn{1}{c}{$n_{2}$} & \multicolumn{1}{c}{$n_{3}$} & \multicolumn{1}{c}{$n_{4}$} & \multicolumn{1}{c}{$n_{5}$} & \multicolumn{1}{c}{$n_{6}$} &\multicolumn{1}{c}{$n_{2}$} & \multicolumn{1}{c}{$n_{3}$} & \multicolumn{1}{c}{$n_{4}$} & \multicolumn{1}{c}{$n_{5}$} & \multicolumn{1}{c}{$n_{6}$}  \\
    \midrule
    \multirow{7}{*}{\rotatebox[origin=c]{90}{\bf Depth}} & FLUX & 44.6 & 18.7 & 6.1 & 1.0 & 0.0 & 62.1 & 50.6 & 42.3 & 39.0 & 29.0 \\
                                     & 3DIS & 63.1 & 38.9 & 25.6 & 13.0 & 16.1 & 77.7 & 71.9 & 62.6 & 55.2 & 54.7 \\
                                     & Regional-FLUX & 59.2 & 35.8 & 18.3 & 9.4 & 7.1 & 74.4 & 68.8 & 62.1 & 57.5 & 50.9 \\
                                     & NoiseCollage & \textbf{66.8} & \textbf{44.5} & \textbf{37.8} & \underline{21.9} & 16.4 & \textbf{79.6} & \textbf{74.8} & \textbf{73.3} & \underline{66.9} & 60.5 \\
                                     & GrounDit & 61.4 & 41.1 & 28.5 & 16.1 & \underline{17.9} & 76.5 & 71.7 & 68.3 & 63.6 & \textbf{62.3} \\
                                     & DreamRenderer & 63.7 & 40.2 & 24.7 & 8.9 & 12.5 & 77.8 & 71.7 & 64.9 & 62.2 & 58.7 \\
                                     & Ours & \underline{65.9} & \underline{42.6} & \underline{32.8} & \textbf{22.9} & \textbf{20.7} & \underline{79.0} & \underline{73.3} & \underline{69.5} & \textbf{67.7} & \underline{62.1} \\
    \midrule
    \multirow{7}{*}{\rotatebox[origin=c]{90}{\bf Canny}} & FLUX & 18.4 & 5.0 & 0.6 & 0.0 & 0.0 & 37.9 & 34.7 & 25.6 & 25.3 & 20.1 \\
                                     & 3DIS & 26.7 & 14.6 & 6.7 & 2.1 & \underline{5.7} & 49.6 & 48.3 & 38.0 & 34.5 & 36.5 \\
                                     & Regional-FLUX & 28.8 & 11.0 & 3.5 & \textbf{3.6} & 0.7 & 49.8 & 46.1 & 40.0 & 39.5 & 35.5 \\
                                     & NoiseCollage & 25.1 & \underline{17.3} & 9.0 & 1.0 & 1.1 & 47.5 & \underline{50.3} & 42.6 & 40.4 & 35.4 \\
                                     & GrounDit & \underline{31.0} & 16.1 & \underline{9.9} & 2.1 & \textbf{7.9} & 49.9 & 49.6 & \underline{47.0} & \underline{41.4} & \underline{44.4} \\
                                     & DreamRenderer & 29.5 & 14.2 & 8.7 & \underline{2.6} & 2.1 & \underline{51.5} & 49.8 & 43.5 & 39.0 & 39.0 \\
                                     & Ours & \textbf{34.6} & \textbf{22.9} & \textbf{11.3} & \underline{2.6} & 5.0 & \textbf{56.1} & \textbf{55.7} & \textbf{50.7} & \textbf{43.9} & \textbf{48.1} \\
    \bottomrule
  \end{tabular}
\end{table}

\subsection{Qualitative Comparison}
\Cref{fig3} illustrates the qualitative comparison between our COLLAR and several state-of-the-art baselines under various spatial conditions. While the vanilla FLUX model generates high-quality images, it often struggles to follow spatial layouts.  Methods such as 3DIS and DreamRenderer occasionally lead to semantic leakage, especially in controlling small-scale instances.  Furthermore, Regional-FLUX and NoiseCollage lack explicit bidirectional interaction between local and global features. This may contribute to visual artifacts in complex scenes. In contrast, our method demonstrates superior spatial fidelity and object-level precision. As shown in the results, generated instances more closely match the input descriptions, with clearer boundaries in overlapping regions. COLLAR also helps preserve instance attributes and overall image realism. The higher visual quality of small objects further validates our model's robust control capability.  In summary, these qualitative results confirm that our approach effectively balances high-fidelity image synthesis with rigorous object-level controllability, outperforming existing methods in complex multi-instance generation.

\subsection{Ablation Study}
\label{ab}
\Cref{tab3} compares component configurations with $K=50$. CSSA with Object-level Feature Injection (OFI) improves spatial control over FLUX. Adding Background Feature Injection (BFI) improves FID and HPSv3, with a small decrease in ISR. The full CFI configuration attains the highest spatial scores in this comparison, but uses a larger maximum injection window ($S=10$) than OFI and OFI+BFI ($S=5$). Its improvement therefore reflects both the component configuration and the guidance schedule. At the same $S=5$ limit, \Cref{tabA3} reports lower ISR/mAP for CFI (42.9/71.4) than OFI (44.6/72.2), alongside better FID (15.6 vs.\ 16.6) and HPSv3 (6.27 vs.\ 6.14). These results demonstrate a trade-off between spatial control and image quality. Additional hyperparameter ablations appear in \Cref{Ablation}.

\noindent \textbf{Inference Efficiency Analysis}. We evaluate 35 depth-conditioned COCO-MIG samples with six objects each on an NVIDIA A800 GPU. \Cref{tab4} reports latency and peak VRAM. COLLAR takes 19.01\,s, compared with 25.15\,s for Regional-FLUX, reducing latency by 24.4\%, while using more memory (35.57 vs.\ 33.89\,GiB). It is also faster than NoiseCollage and DreamRenderer, but slower than FLUX and 3DIS. Early stopping reduces COLLAR's latency from 30.23 to 19.01\,s (37.1\%) without reducing peak VRAM. These measurements characterize the computation cost alongside the quality and control results in \Cref{tab1}.

\begin{table}[tb]
  \caption{Component configurations at $K=50$. OFI and OFI+BFI use $S=5$; full CFI uses $S=10$.}
  \label{tab3}
  \centering
  \scriptsize
  \begin{tabular}{@{}llccccccccc@{}}
    \toprule
    \multicolumn{4}{c}{} & \multicolumn{3}{c}{OSR$\uparrow$} & \multicolumn{4}{c}{} \\
    Input & Method & L-CLIP$\uparrow$ & ISR$\uparrow$ & \multicolumn{1}{c}{$n_{3}$} & \multicolumn{1}{c}{$n_{4}$} &\multicolumn{1}{c}{$n_{5}$}  & mIoU$\uparrow$  & mAP$\uparrow$ & FID$\downarrow$ & HPSv3$\uparrow$ \\
    \midrule
    \multirow{4}{*}{\rotatebox[origin=c]{90}{\bf Depth}} & FLUX & 26.97 & 24.1 & 50.6 & 42.3 & 39.0 & 43.7 & 47.5 & /  & 6.14 \\
                                     & FLUX + CSSA + OFI  & 28.1 & 44.6 & 73.6 & 67.7 & 64.5 & 64.1 & 72.2 & 16.6 & 6.14 \\
                                     & FLUX + CSSA + OFI + BFI & 28.1 & 44.2 & 73.4 & 67.7 & 64.2 & 64.2 & 72.2 & 16.2 & 6.22 \\
                                     & FLUX + CSSA + CFI   & \textbf{28.2} & \textbf{45.3} & \textbf{73.8} & \textbf{68.3} & \textbf{65.6} & \textbf{64.5} & \textbf{73.1} & 16.8 & 6.17 \\
    \bottomrule
  \end{tabular}
\end{table}

\begin{table}[tbhp!]
	\caption{Computational efficiency comparison (1024$\times$1024). ES: early-stopping.}
	\centering
	  \scriptsize
	\begin{tabular}{@{}lcccccccc@{}}
		\toprule  
		\textbf{Methods}&FLUX&3DIS&Regional-FLUX&GrounDit&NoiseCollage&DreamRenderer&Ours(w/o ES)&Ours\\ 
		\cmidrule(r){1-9}
		\textbf{Inference Time(s)}& 10.12 & 16.83 & 25.15 & 19.61 & 39.81 & 36.71 & 30.23 & 19.01 \\
		\textbf{Peak VRAM(GiB)}& 33.82 & 33.83 & 33.89 & 33.85 & 33.86 & 33.84 & 35.57 & 35.57 \\
		\bottomrule  
	\end{tabular}
	
\label{tab4}
\end{table}

\FloatBarrier
\subsection{Integration with other FLUX-based Conditional Methods}
\label{t1}
\begin{figure}[htbp!]
  \centering
   \includegraphics[height=8cm]{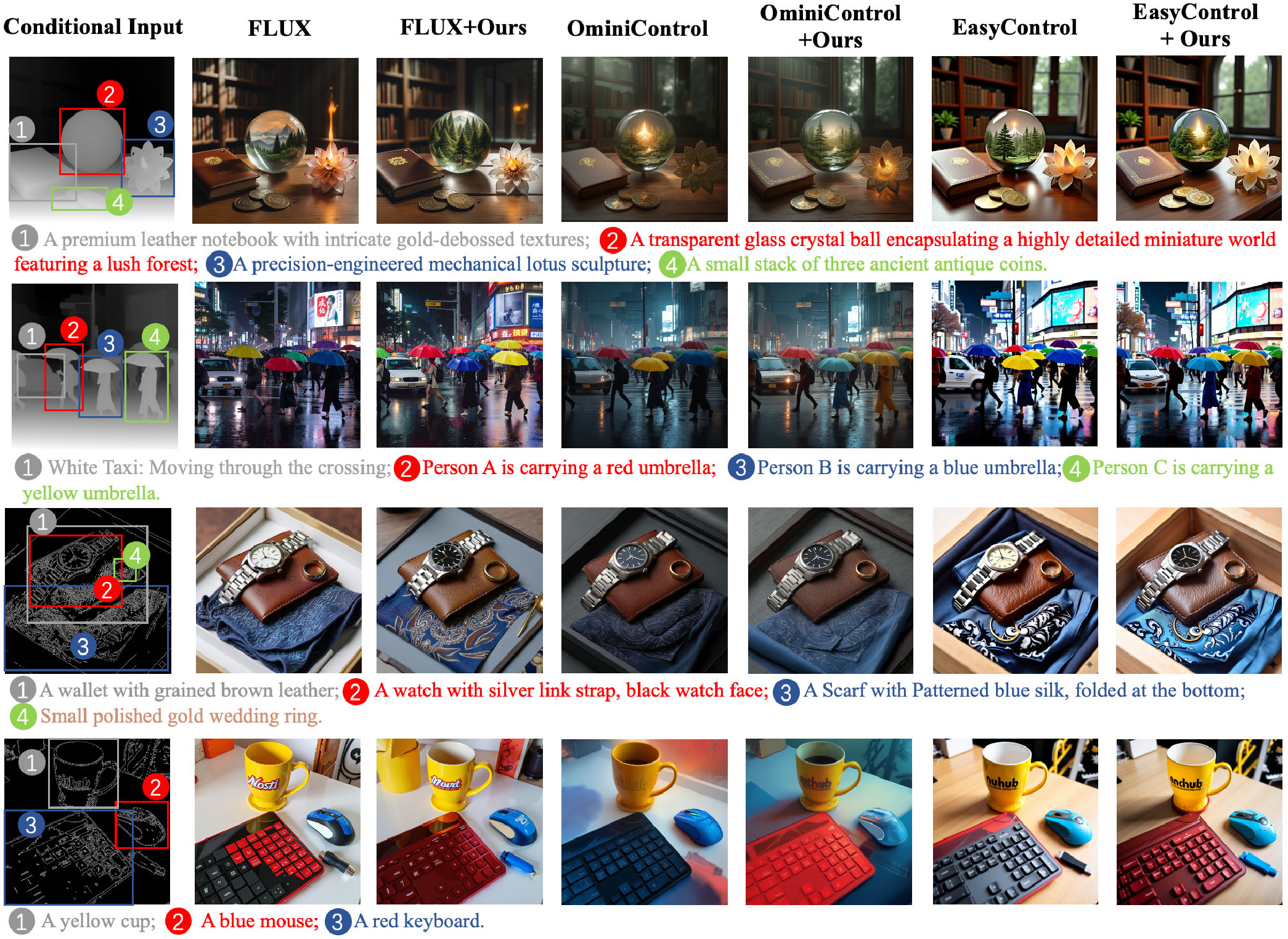}
   \caption{Qualitative results demonstrating the generalizability of our method as a plug-and-play module.}
   \label{figA3}
\end{figure}

\Cref{tab6} demonstrates the compatibility and effectiveness of our proposed method when integrated with the state-of-the-art conditional generation frameworks OminiControl~\cite{c60} and EasyControl~\cite{c6}.  The quantitative results reveal that our cascaded refinement strategy improves the object-level controllability of both baselines, particularly in instance rendering and spatial alignment.  Notably, when applied to OminiControl, our method yields a substantial increase in ISR (from 18.62 to 27.00) and mAP (from 44.70 to 57.77).  The performance gains are even more pronounced for EasyControl, where integrating our strategy increases the OSR from 47.61 to 64.93 and increases the ISR from 25.46 to 36.83.  These results show that COLLAR improves object-level control on both backbones, with comparable G-CLIP and higher L-CLIP scores.  Such quantitative leaps are further corroborated by the qualitative visualizations (\Cref {figA3}), which clearly illustrate our method's superior ability to harmonize complex multi-object layouts and preserve geometric fidelity compared to the original baselines.

\begin{table}[htbp!]
  \caption{Integration with FLUX-based conditional generation methods on COCO-MIG. Within each backbone pair, \textbf{bold} denotes the better result.}
  \label{tab6}
  \centering
  \small
  \begin{tabular}{@{}llcccccc@{}}
    \toprule
    Input & Method & G-CLIP$\uparrow$ & L-CLIP$\uparrow$ & ISR$\uparrow$ & OSR$\uparrow$ & mIoU$\uparrow$  & mAP$\uparrow$  \\
    \midrule
    \multirow{4}{*}{\rotatebox[origin=c]{90}{\bf Depth}}
                                     & OminiControl  & 31.22 & 27.25 & 18.62 & 43.50 & 41.78 & 44.70 \\
                                     & OminiControl + Ours  & \textbf{31.30} & \textbf{27.88} & \textbf{27.00} & \textbf{56.02} & \textbf{53.16} & \textbf{57.77} \\
     
                                     & EasyControl  & \textbf{30.97} & 26.24 & 25.46 & 47.61 & 45.25 & 48.54 \\
                                     & EasyControl + Ours  & 30.94 & \textbf{27.34} & \textbf{36.83} & \textbf{64.93} & \textbf{59.78} & \textbf{66.37} \\
    \bottomrule
  \end{tabular}
\end{table}

\FloatBarrier
\section{Conclusion}
\label{conclusion}
We present COLLAR, a training-free, plug-and-play framework for precise instance-level control in spatially-conditioned image generation. It introduces two core modules: (1) Cross-Scale Semantic Alignment, which leverages an extended Field-of-View branch to seamlessly bridge local semantics with global spatial contexts; and (2) Cyclic Feature Injection, which employs a frequency-based adaptive strategy for the seamless blending of local details into the global backbone. Experiments show that COLLAR achieves competitive performance and can be integrated into different generative models.

\textbf{Limitation.} COLLAR's local and extended-FoV branches increase computation relative to FLUX and 3DIS. Although it runs faster than NoiseCollage and DreamRenderer in our comparison, its peak memory usage is higher. Lightweight latent fusion or inference distillation may reduce this cost. Resolving conflicts between local textual semantics and spatial geometry also remains challenging; a semantic-structural-aware feature-scaling strategy is a direction for future work.

\section*{Acknowledgments}
We would like to thank the reviewers for their thoughtful comments. The work is supported in part by the Zhejiang “Leading Goose” R\&D Program (No. 2025C02211) and the Ningbo “Yongjiang Science Innovation 2035” Key Technology Breakthrough Plan (Nos. 2024Z254 and 2025Z056).

\bibliographystyle{acl_natbib}
\bibliography{main}

\clearpage
\appendix
\section*{Appendix}

\section{User Study}
\label{userstudy}
\begin{figure}[htbp!]
  \centering
  \begin{subfigure}[b]{0.49\linewidth}
   \centering
    \includegraphics[height=4.5cm]{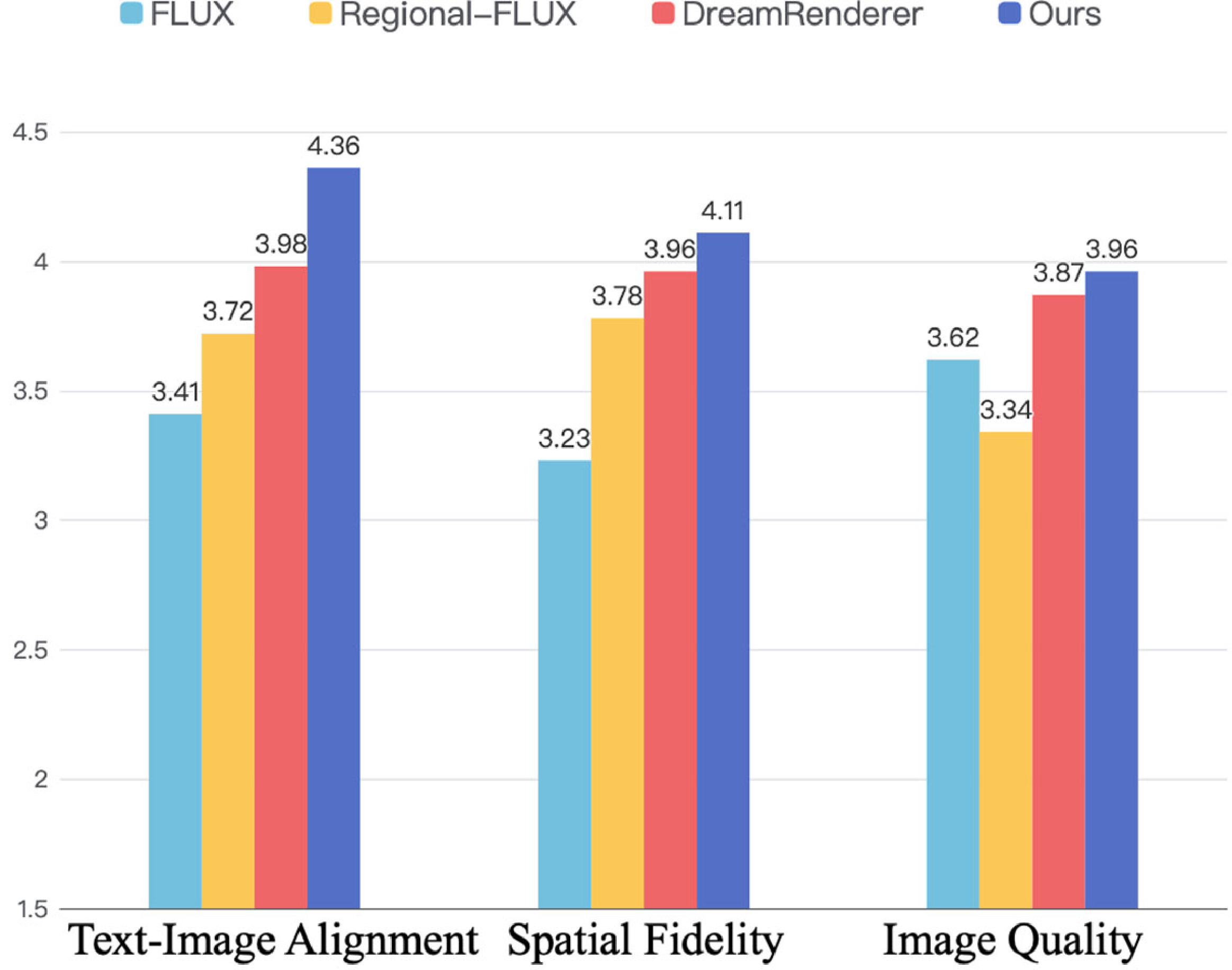} 
    \caption{Users were asked to rate: text-image alignment, spatial fidelity, and image quality.}
    \label{fig:short-a}
  \end{subfigure}
  \hfill
  \begin{subfigure}[b]{0.49\linewidth}
  \centering
    \includegraphics[height=4.5cm]{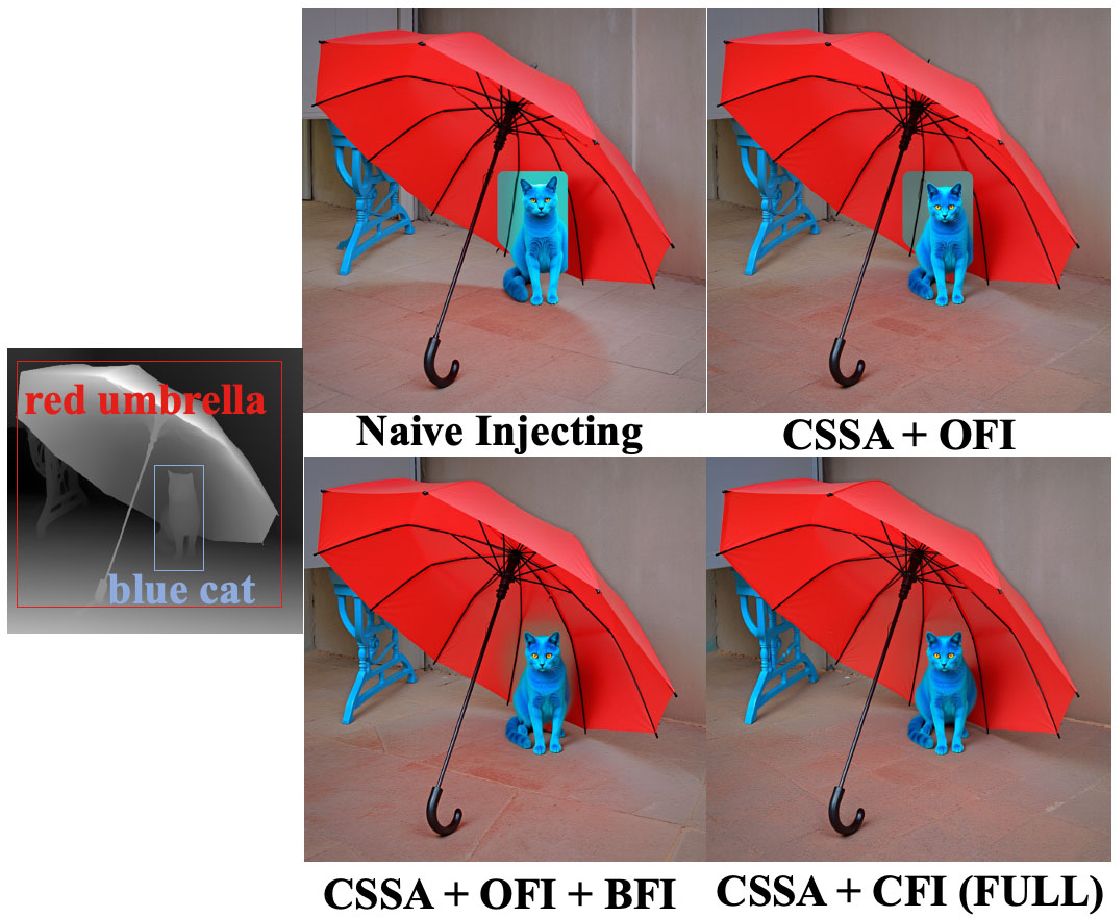} 
    \caption{Ablation study of model designs. Our method can achieve spatial fidelity and image quality.}
    \label{fig:short-b}
  \end{subfigure}
  \caption{Visualization results of User Study and Ablation Study.}
  \label{fig4}
\end{figure}

\begin{figure}[htbp!]
  \centering
   \includegraphics[height=8cm]{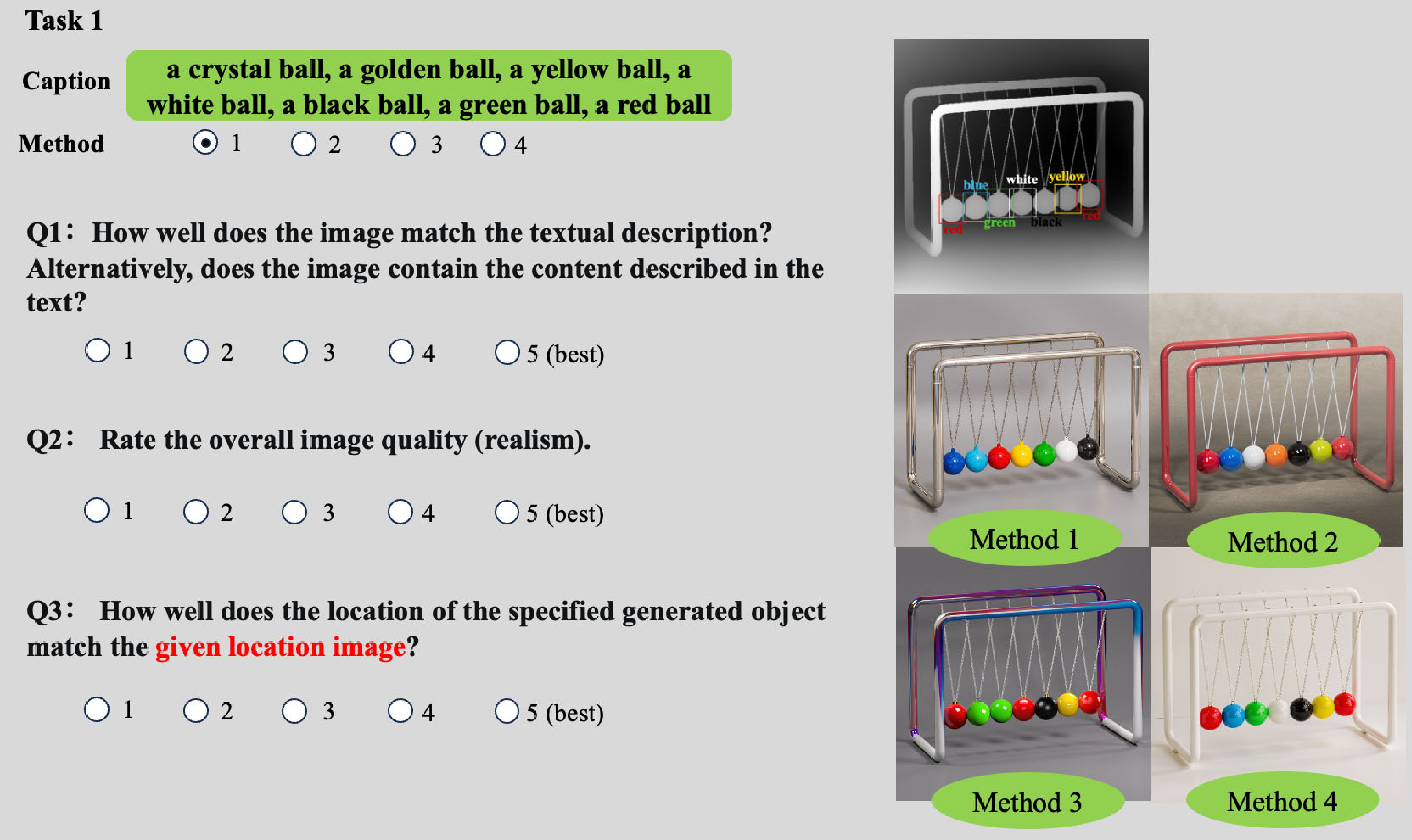}
   \caption{Screenshot of the instruction page and evaluation interface provided to participants during the user study.}
   \label{figscreenshot}
\end{figure}

\Cref{fig4} (a) shows a user study with 20 participants with a background in computer science, comprising 12 Master's students (60\%) and 8 Ph.D. candidates (40\%). Four methods are selected for evaluation: COLLAR, DreamRenderer, Regional-FLUX, and FLUX. We randomly selected 40 generated images from the COCO-MIG benchmark for each participant and asked them to rate each image on text-image alignment, image quality, and spatial fidelity. The rating scale ranged from 1 to 5 (highest). Fig.~\ref{fig4}(a) shows that users preferred our method for spatial fidelity and image quality. A screenshot of the evaluation interface and the instructions provided to participants are shown in \Cref{figscreenshot}.

\section{Ablation Study}
\label{Ablation}

\subsection{Validation of the High-Pass Residual}
\label{sec:highpass-validation}
We implement $\mathcal{H}(z)=z-\operatorname{AvgPool}_{3\times3}(z)$ using stride 1 and padding 1. Local averaging yields a low-pass-like component, while subtraction yields a high-pass-like residual. We examine their frequency content using channel-wise spatial 2-D FFTs of the final latents. Let $r$ denote radial frequency normalized by the corner Nyquist radius. We define $r\leq0.25$ as the low band and $r\geq0.50$ as the high band. Each energy fraction is normalized by the corresponding component's total power. The middle band is omitted, so the two fractions need not sum to 100\%.

\begin{table}[htbp!]
  \caption{Frequency-band energy fractions (\%). Avg.: average-pooled component; Res.: residual.}
  \label{tab:frequency-energy}
  \centering
  \small
  \begin{tabular}{@{}llrrrr@{}}
    \toprule
    Condition & Kernel & Avg. low & Avg. high & Res. low & Res. high \\
    \midrule
    Depth & $3\times3$ & 96.39 & 0.79 & 2.17 & 74.20 \\
    Depth & $5\times5$ & 98.64 & 0.39 & 9.94 & 48.00 \\
    Canny & $3\times3$ & 95.59 & 0.93 & 2.49 & 72.60 \\
    Canny & $5\times5$ & 98.28 & 0.49 & 11.12 & 45.32 \\
    \bottomrule
  \end{tabular}
\end{table}

\Cref{tab:frequency-energy} shows that local averaging concentrates energy in the low-frequency band. Compared with the $5\times5$ kernel, the $3\times3$ kernel produces a residual with a higher proportion of high-frequency energy and a lower proportion of low-frequency energy under both Depth and Canny conditioning. These results support our choice of the $3\times3$ kernel for extracting the high-pass residual.

We then examine whether the residual captures contour information after decoding. At the final denoising step, we decode the average-pooled component and the residual separately. The image decoded from the full latent serves as the reference. For Pixel Edge AP~\cite{c61}, we compare Scharr gradient magnitudes in each component image with Canny edges from the reference image. Boundary F1~\cite{c62} measures contour agreement with the reference image. As shown in \Cref{tab:edge-response}, the residual achieves higher scores on both metrics under Depth and Canny conditioning.

\begin{table}[htbp!]
  \caption{Contour responses of decoded latent components.}
  \label{tab:edge-response}
  \centering
  \small
  \begin{tabular}{@{}llcc@{}}
    \toprule
    Condition & Component & Pixel Edge AP$\uparrow$ & Boundary F1$\uparrow$ \\
    \midrule
    Depth & Average-pooled & 0.06 & 0.23 \\
          & High-pass residual & \textbf{0.32} & \textbf{0.73} \\
    Canny & Average-pooled & 0.09 & 0.34 \\
          & High-pass residual & \textbf{0.48} & \textbf{0.83} \\
    \bottomrule
  \end{tabular}
\end{table}

\subsection{Effect of the Early-Stopping Signal}
\label{sec:stopping-signal}
We compare full-latent and high-pass-residual signals for early stopping using the same independent set of 600 test samples as the object-size evaluation in \Cref{sec:object-size-evaluation}. \Cref{tab:stopping-signal} reports spatial control, image quality, and mean injection steps. The high-pass signal reduces mean injection steps from 8.25 to 7.29 and improves HPSv3 and PickScore, with decreases of only 0.13 and 0.31 percentage points in OSR and mIoU, respectively.

\begin{table}[htbp!]
  \caption{Early-stopping signal comparison on the independent 600-sample test set under Depth conditioning.}
  \label{tab:stopping-signal}
  \centering
  \small
  \begin{tabular}{@{}lccccc@{}}
    \toprule
    Signal & OSR$\uparrow$ & mIoU$\uparrow$ & HPSv3$\uparrow$ & PickScore$\uparrow$ & Mean steps$\downarrow$ \\
    \midrule
    Full latent & \textbf{69.77} & \textbf{64.46} & 5.97 & 0.49 & 8.25 \\
    High-pass & 69.64 & 64.15 & \textbf{6.19} & \textbf{0.51} & \textbf{7.29} \\
    \bottomrule
  \end{tabular}
\end{table}

We further group objects by their bounding-box area relative to image area, denoted by $a$. The stop rate is the percentage of objects that meet the similarity-based stopping criterion before the maximum injection step $S=10$. \Cref{tab:stopping-size} shows higher stop rates for the high-pass signal in all three size ranges, with larger objects stopping more frequently.

\begin{table}[htbp!]
  \caption{Early-stop rates (\%) by relative object area $a$ under Depth conditioning.}
  \label{tab:stopping-size}
  \centering
  \small
  \begin{tabular}{@{}lccc@{}}
    \toprule
    Signal & $a<5\%$ & $5\%\leq a<15\%$ & $a\geq15\%$ \\
    \midrule
    Full latent & 14.20 & 34.01 & 51.14 \\
    High-pass & \textbf{15.34} & \textbf{41.57} & \textbf{70.83} \\
    \bottomrule
  \end{tabular}
\end{table}

\subsection{Hyperparameter Ablations}
\noindent \textbf{Effect of FoV Expansion Factor $\gamma$}.
The parameter $\gamma$ determines the proportion of the extended field-of-view in the CSSA module. In \Cref{tabA3}, as $\gamma$ increases from $0.2$ to $1.0$, we observe a consistent reduction in FID ($16.2 \rightarrow 15.3$). This improvement suggests that incorporating more surrounding spatial context through a larger FoV expansion may help preserve image quality. However, an excessively large FoV tends to relax the local spatial constraints, resulting in a slight decrease in ISR ($43.4 \rightarrow 42.3$) and mIoU ($63.4 \rightarrow 62.8$). We adopt $\gamma=0.5$ as an empirical compromise between image quality and spatial fidelity.

\begin{table}[htbp!]
  \caption{Ablation study on model designs and hyperparameters (under the injecting layers $K$ =50 ).}
  \label{tabA3}
  \centering
  \scriptsize
  \begin{tabular}{@{}llccccccccc@{}} 
    \toprule
    \multicolumn{4}{c}{} & \multicolumn{3}{c}{OSR$\uparrow$} & \multicolumn{4}{c}{} \\
    Input & Method & L-CLIP$\uparrow$ & ISR$\uparrow$ & \multicolumn{1}{c}{$n_{3}$} & \multicolumn{1}{c}{$n_{4}$} &\multicolumn{1}{c}{$n_{5}$}  & mIoU$\uparrow$  & mAP$\uparrow$ & FID$\downarrow$  & HPSv3$\uparrow$  \\
    \midrule
    \multirow{11}{*}{\rotatebox[origin=c]{90}{\bf Depth}} & FLUX & 26.97 & 24.1 & 50.6 & 42.3 & 39.0 & 43.7 & 47.5 & / & 6.14 \\
                                     & FLUX + CSSA + OFI  &  & &  &  &  &  & & &  \\
                                     & $\bullet$ $\gamma=0.5, S=5$  & 28.1 & 44.6 & 73.6 & 67.7 & 64.5 & 64.1 & 72.2 & 16.6 & 6.14 \\
                                     & FLUX + CSSA + OFI + BFI  &  & &  &  &  &  & & &  \\
                                     & $\bullet$ $\gamma=0.5, S=5$ & 28.1 & 44.2 & 73.4 & 67.7 & 64.2 & 64.2 & 72.2 & 16.2 & 6.22 \\
                                     & FLUX + CSSA + CFI  &  & &  &  &  &  & & &   \\
                                     & $\bullet$ $\gamma=0.2, S=5, \sigma=0.85$  & 28.1 & 43.4 & 73.1 & 67.2 & 64.3 & 63.4 & 71.6 & 16.2 & 6.20 \\
                                     & $\bullet$ $\gamma=0.5, S=5, \sigma=0.85$  & 28.1 & 42.9 & 72.8 & 66.5 & 63.6 & 63.3 & 71.4 & 15.6 & 6.27 \\
                                     & $\bullet$ $\gamma=1.0, S=5, \sigma=0.85$   & 28.1 & 42.3 & 72.6 & 65.0 & 62.6 & 62.8 & 70.9 & \textbf{15.3} & \textbf{6.31} \\
                                     & $\bullet$ $\gamma=0.5, S=10, \sigma=0.80$    & 28.1 & 41.7 & 72.5 & 65.0 & 62.5 & 62.7 & 70.7 & 15.8 & 6.25 \\
                                     & $\bullet$ $\gamma=0.5, S=10, \sigma=0.85$ & \textbf{28.2} & \textbf{45.3} & \textbf{73.8} & \textbf{68.3} & \textbf{65.6} & \textbf{64.5} & \textbf{73.1} & 16.8 & 6.17 \\
    \bottomrule
  \end{tabular}
\end{table}

\noindent \textbf{Effect of Guidance Step Control ($\sigma$ and $S$)}.
The parameters $S$ and $\sigma$ jointly determine when feature injection terminates. $S$ sets the maximum number of guidance steps, while $\sigma$ defines the similarity threshold for early stopping. With $\gamma=0.5$ and $\sigma=0.85$, increasing $S$ from $5$ to $10$ improves mAP from $71.4$ to $73.1$ and ISR from $42.9$ to $45.3$. At $S=10$, increasing $\sigma$ from $0.80$ to $0.85$ improves mIoU from $62.7$ to $64.5$. These results suggest that permitting longer feature injection can improve spatial fidelity. However, the corresponding increase in FID and decrease in HPSv3 suggest a potential trade-off between spatial control and image quality.

\noindent \textbf{Effect of the Number of Injecting Layers $K$ and Weight $\lambda$}. \Cref{tab5} provides a comprehensive quantitative analysis of the number of injecting layers $K$ and the balance factor $\lambda$, revealing a clear performance scaling effect. Other hyperparameters follow the selected configuration presented in \cref{tabA3}. As $K$ increases from 20 to 57, spatial accuracy metrics such as mIoU and mAP exhibit an overall upward trend, indicating that deeper feature injection is pivotal for reinforcing geometric constraints within the DiT architecture. Increasing $K$ from 50 to 57 improves mIoU and mAP while reducing FID from 16.8 to 16.3. With $K=57$, setting $\lambda=0.5$ increases ISR to 47.2 and mAP to 74.9. This configuration also demonstrates enhanced robustness in complex multi-object scenarios, with the OSR for $n_5$ rising from 67.7 to 69.7, indicating improved instance-level control. Overall, we select $K=57$ and $\lambda=0$ for the main experiments because this configuration provides a better balance between image quality and spatial control. Compared with $\lambda=0.5$, it reduces FID from 17.1 to 16.3 while retaining competitive instance-level control.

\begin{table}[htbp!]
  \caption{Ablation study on the number of injecting layers $K$ and weight $\lambda$.}
  \label{tab5}
  \centering
  \scriptsize
  \begin{tabular}{@{}llcccccccc@{}}
    \toprule
    \multicolumn{4}{c}{} & \multicolumn{3}{c}{OSR$\uparrow$} & \multicolumn{3}{c}{} \\
    Input & Method & L-CLIP$\uparrow$ & ISR$\uparrow$ & \multicolumn{1}{c}{$n_{3}$} & \multicolumn{1}{c}{$n_{4}$} &\multicolumn{1}{c}{$n_{5}$}  & mIoU$\uparrow$  & mAP$\uparrow$ & FID$\downarrow$  \\
    \midrule
    \multirow{7}{*}{\rotatebox[origin=c]{90}{\bf Depth}}
                                     & FLUX + CSSA + CFI  &  & &  &  &  &  & &  \\
                                     & $\bullet$ $K=20, \lambda =0 $  & 28.05 & 43.0 & 73.3 & 67.2 & 62.2 & 63.5 & 71.6 & \textbf{15.9} \\
                                     & $\bullet$ $K=30, \lambda =0$  & 28.08 & 43.3 & 72.4 & 66.4 & 64.0 & 63.7 & 71.7 & 16.6 \\
                                     & $\bullet$ $K=40, \lambda =0$   & 28.15 & 44.3 & 73.6 & 68.0 & 65.1 & 64.3 & 72.7 & 16.7 \\
                                     & $\bullet$ $K=50, \lambda =0$    & 28.2 & 45.3 & 73.8 & 68.3 & 65.6 & 64.5 & 73.1 & 16.8 \\
                                     & $\bullet$ $K=57, \lambda =0$ (Ours)  & 28.16 & 46.8 & 73.3 & 69.5 & 67.7 & 65.0 & 73.8 & 16.3 \\
                                     & $\bullet$ $K=57, \lambda =0.5$   & \textbf{28.23} & \textbf{47.2} & \textbf{74.4} & \textbf{69.9} & \textbf{69.7} & \textbf{65.6} & \textbf{74.9} & 17.1 \\
    \bottomrule
  \end{tabular}
\end{table}

\subsection{Effect of Object Count on Inference Cost}
\label{sec:object-count-cost}
We evaluate computational cost using 35 samples for each object count and report mean inference time and mean peak GPU memory in \Cref{tab:object-count-cost}. We use the same inference implementation as in \Cref{tab4}. As the number of objects increases from two to six, mean inference time increases from 14.05 to 19.01 seconds, while mean peak GPU memory increases from 34.96 to 35.57 GiB. The measurements show approximately linear runtime growth over the tested range.

\begin{table}[htbp!]
  \caption{Inference cost for different numbers of objects. Each column averages 35 samples.}
  \label{tab:object-count-cost}
  \centering
  \small
  \begin{tabular}{@{}lccccc@{}}
    \toprule
    Number of objects & 2 & 3 & 4 & 5 & 6 \\
    \midrule
    Mean inference time (s) & 14.05 & 15.29 & 16.31 & 17.70 & 19.01 \\
    Mean peak GPU memory (GiB) & 34.96 & 35.10 & 35.19 & 35.28 & 35.57 \\
    \bottomrule
  \end{tabular}
\end{table}

\section{Integration with SD3-based Conditional Methods}
\label{t2}
To demonstrate that our method can be generalized to other DiT-based models, \Cref{tab7} presents a quantitative comparison utilizing Stable Diffusion 3 (SD3) equipped with ControlNet as the backbone. The results clearly indicate that our method outperforms both the vanilla baseline and DreamRenderer in precise spatial control. Specifically, our approach achieves substantial improvements across all primary layout-alignment metrics, yielding the highest ISR (27.92), OSR (57.25), and mIoU (52.73). Furthermore, our method secures the best local semantic alignment. While the vanilla SD3+ControlNet retains slightly higher global metrics (e.g., G-CLIP and PickScore)—which is typical for native models generating relatively unconstrained scenes—it shows limited multi-instance spatial adherence. In contrast, our method successfully bridges this gap, achieving the best object-level controllability among the compared methods on the SD3 architecture while maintaining highly competitive overall aesthetic quality.

\begin{table}[htbp!]
  \caption{Quantitative comparisons on COCO-MIG using SD3 with ControlNet. \textbf{Bold} and \underline{underline} denote the best and second-best reported results, respectively.}
  \label{tab7}
  \centering
  \scriptsize
  \begin{tabular}{@{}llccccccccc@{}}
    \toprule
    Input & Method & G-CLIP$\uparrow$ & L-CLIP$\uparrow$ & ISR$\uparrow$ & OSR$\uparrow$ & mIoU$\uparrow$  & mAP$\uparrow$ & FID$\downarrow$ & HPSv3$\uparrow$ & Pick$\uparrow$  \\
    \midrule
    \multirow{3}{*}{\rotatebox[origin=c]{90}{\bf Depth}}
                                     & SD3 + ControlNet  & \textbf{33.22} & 25.68 & 9.96 & 31.57 & 33.43 & 31.25 & / & \textbf{5.502} & \textbf{0.4033} \\
                                     & DreamRenderer  & 32.94 & \underline{26.50} & \underline{17.50} & \underline{46.16} & \underline{44.29} & \underline{47.44} & \underline{21.53} & 3.849 & 0.2429 \\
                                     & Ours  & \underline{33.02} & \textbf{27.15} & \textbf{27.92} & \textbf{57.25} & \textbf{52.73} & \textbf{58.54} & \textbf{18.66} & \underline{5.435} & \underline{0.3537} \\
    \bottomrule
  \end{tabular}
\end{table}

\clearpage
\section{Additional Evaluation}
\label{sec:additional-evaluation}
\subsection{Performance across Object Sizes}
\label{sec:object-size-evaluation}
We evaluate OSR on COCO-MIG across three object-size groups, defined by the ratio $a$ of target bounding-box area to image area: small ($a<5\%$), medium ($5\%\leq a<15\%$), and large ($a\geq15\%$). We use a separate set of 600 randomly selected test samples, also used in the early-stopping analysis in \Cref{sec:stopping-signal}. An object is counted as successful only when both its category and color attribute are correct. \Cref{tab:size-osr} reports the results. With Canny conditioning, COLLAR achieves the highest OSR across all three size groups.

\begin{table}[htbp!]
  \caption{COCO-MIG OSR (\%) by object size.}
  \label{tab:size-osr}
  \centering
  \small
  \setlength{\tabcolsep}{4pt}
  \begin{tabular}{@{}llccc@{}}
    \toprule
    Condition & Method & Small ($<5\%$) & Medium (5--15\%) & Large ($\geq15\%$) \\
    \midrule
    Depth & NoiseCollage & 65.9 & \underline{73.6} & \textbf{73.2} \\
    & GrounDiT & \textbf{67.4} & 71.9 & 67.6 \\
    & DreamRenderer & 62.9 & 68.5 & \underline{71.4} \\
    & COLLAR & \underline{66.7} & \textbf{73.7} & 71.0 \\
    \midrule
    Canny & NoiseCollage & 32.9 & 43.4 & \underline{53.0} \\
    & GrounDiT & \underline{42.7} & \underline{48.0} & 49.5 \\
    & DreamRenderer & 36.4 & 45.2 & 52.8 \\
    & COLLAR & \textbf{47.4} & \textbf{52.8} & \textbf{54.0} \\
    \bottomrule
  \end{tabular}
\end{table}

\clearpage
\subsection{Additional Qualitative Results}
\label{sec:additional-qualitative}
\begin{figure}[htbp!]
  \centering
  \includegraphics[height=8cm]{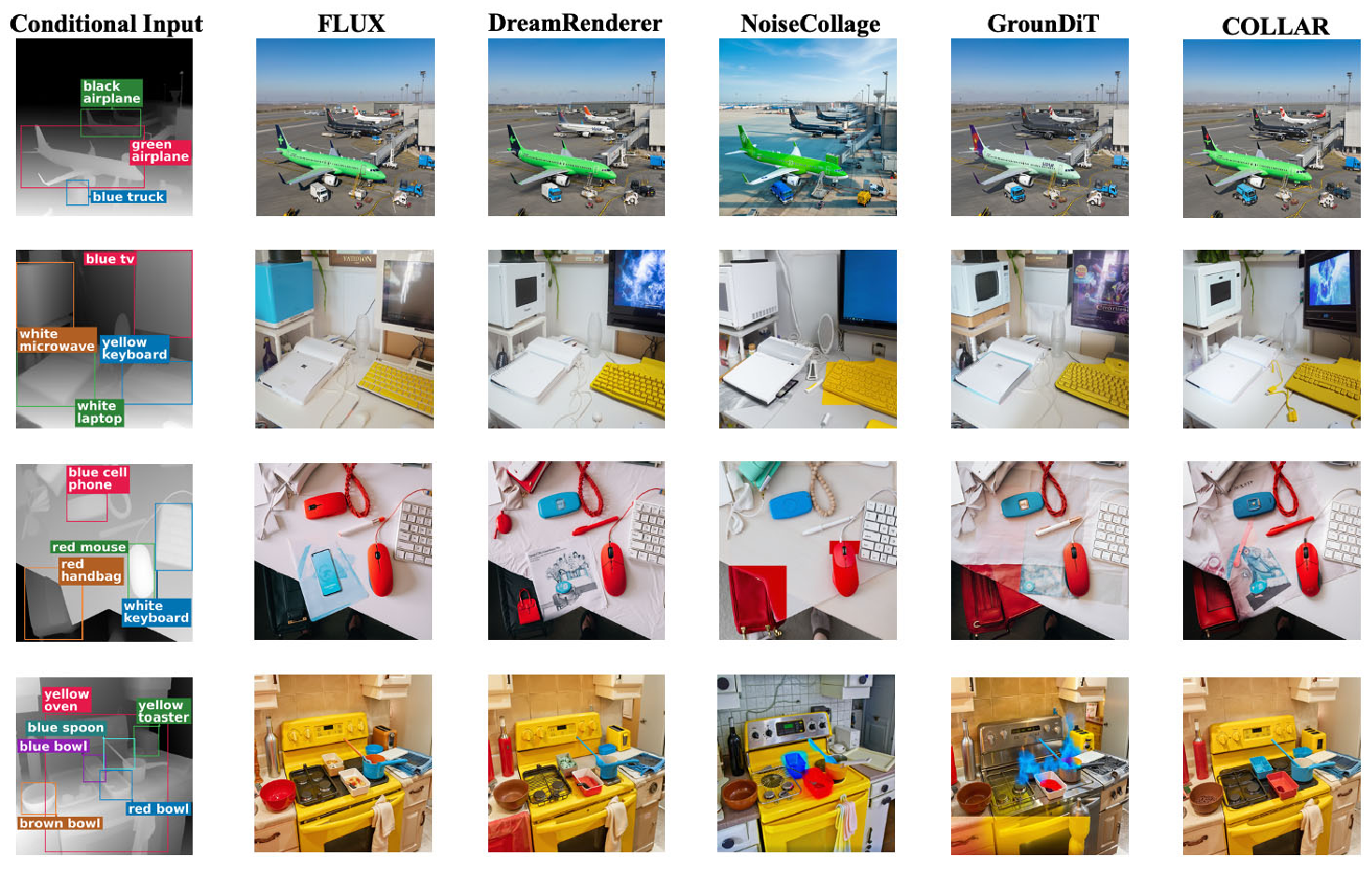}
  \caption{Additional qualitative comparisons under depth conditioning.}
  \label{fig:additional-depth}
\end{figure}

\begin{figure}[htbp!]
  \centering
  \begingroup
  \setlength{\unitlength}{\dimexpr8cm/1316\relax}
  \begin{picture}(2094,1316)
    \put(0,0){\includegraphics[height=8cm]{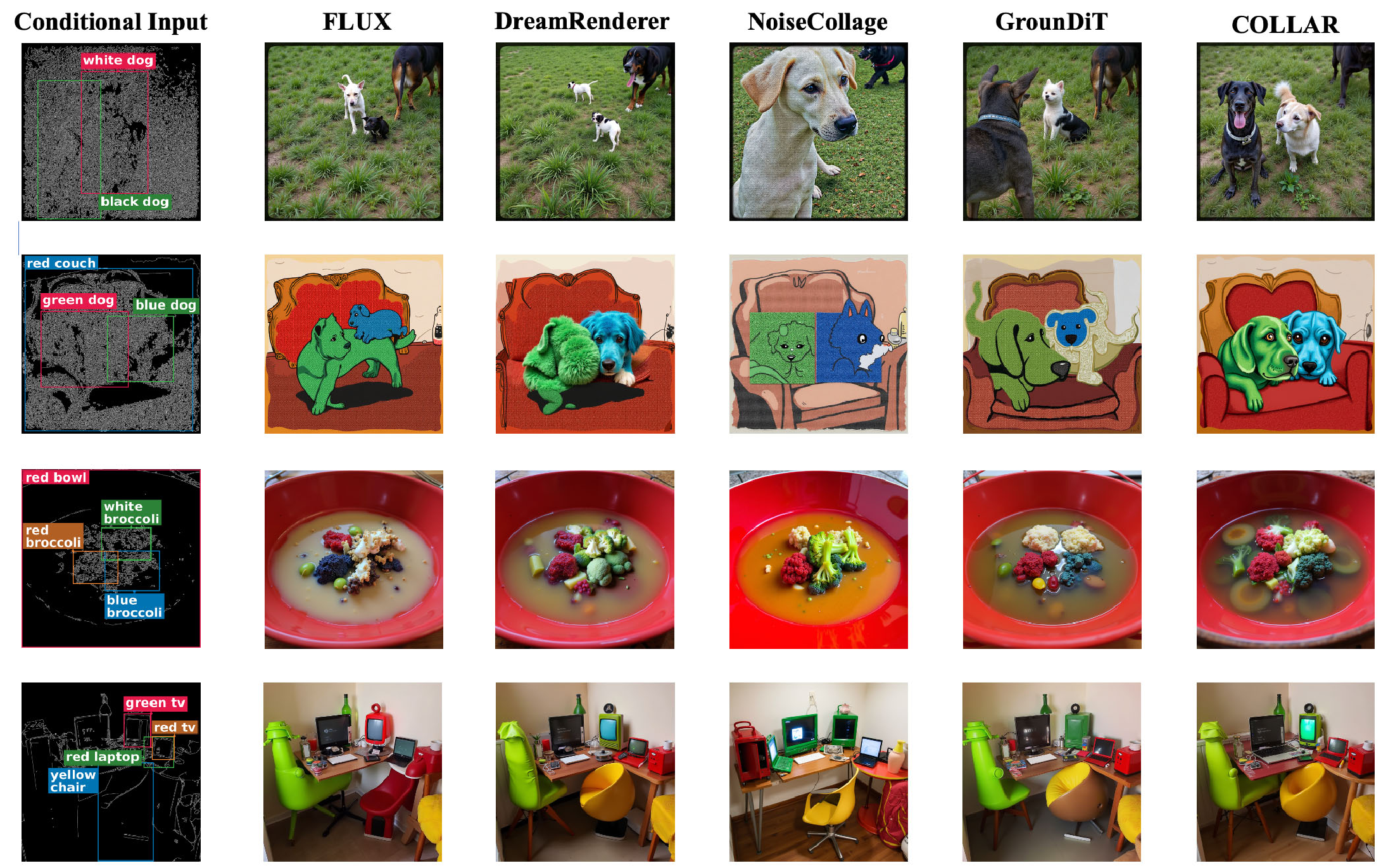}}
    \put(24,925){\color{white}\rule{8\unitlength}{60\unitlength}}
  \end{picture}
  \endgroup
  \caption{Additional qualitative comparisons under Canny conditioning.}
  \label{fig:additional-canny}
\end{figure}

\Cref{fig:additional-depth,fig:additional-canny} show additional qualitative comparisons under depth and Canny conditioning, respectively.

\clearpage
\section{Pseudocode}
\Cref{alg1} illustrates the complete inference process of our method.

\begin{algorithm}[H]
\small
\caption{COLLAR: Cascaded Object-Level Latent Refinement}
\label{alg1}
\begin{algorithmic}[1]
\renewcommand{\algorithmicrequire}{\textbf{Input:}}
\renewcommand{\algorithmicensure}{\textbf{Output:}}
\REQUIRE Global prompt $P$, conditional image input $C$, bounding boxes $\{B_i\}$, object descriptions $\{O_i\}$.
\STATE \textbf{Parameters:} Threshold $\sigma$, maximum guidance steps $S$, FoV expansion factor $\gamma$, weight $\lambda$, high-pass filter $\mathcal{H}(\cdot)$, injection-layer set $\mathcal{K}$.
\STATE \textbf{Initialize:} $z_{G}^{(T)}, \{z_{i, L}^{(T)},\}, \{z_{i, E}^{(T)},\} \sim \mathcal{N}(0, \mathbf{I})$ \COMMENT{Global, Local, and Ex-FoV latents}
\STATE $\mathcal{A} \gets \{1, \dots, N\}$ \COMMENT{Set of active instances for feature injection}
\STATE $C_G \gets C$
\FOR{$t = T, T-1, \dots, 1$}
    \FOR{each instance $i \in \mathcal{A}$}
        \STATE $B_{i,E} \gets \text{Expand}(B_i,\gamma)$ \COMMENT{Extended-FoV boxes}
	\STATE $C_{i,L} \gets \text{Crop}(C,B_i),\quad C_{i,E} \gets \text{Crop}(C,B_{i,E})$ \COMMENT{Branch conditions}
	\STATE $M_{i,E},M_{i,G} \gets \text{BuildMasks}(B_i,B_{i,E})$ \COMMENT{Ex-FoV and globally aligned masks}
        \STATE // \textbf{Step 1: Local Semantic Extraction}
 
        \STATE $\epsilon_{i, L}^{(t)}, \{ (Q, K, V)_{i, L}^{(t, k)} \}_{k \in \mathcal{K}} \gets \text{DiT}_{L}(z_{i, L}^{(t)}, O_i, C_{i,L})$ \COMMENT{Extract QKV from Local branch}
        \STATE $z_{i, L}^{(t-1)} \gets \text{Scheduler}(z_{i, L}^{(t)}, \epsilon_{i, L}^{(t)}, t)$
	\STATE // \textbf{Step 2: Global Background Latent Injecting}
        \STATE $z_{i, E}^{(t)} \gets \text{Fuse}(z_{i, E}^{(t)}, z_{G}^{(t)}, [B_{i, E}])$ \COMMENT{Inject global background into Ex-FoV latent}
        
        \STATE // \textbf{Step 3: Extended-FoV Feature Fusion}
        \STATE $\epsilon_{i, E}^{(t)},  \{ f_{i, E}^{(t, k)} \}_{k \in \mathcal{K}} \gets \text{DiT}_{E}(z_{i, E}^{(t)},\{ (Q, K, V)_{i, L}^{(t, k)} \}_{k \in \mathcal{K}}, O_i, C_{i,E})$ 
         \STATE $z_{i, E}^{(t-1)} \gets \text{Scheduler}(z_{i, E}^{(t)}, \epsilon_{i, E}^{(t)}, t)$
        \STATE // \textbf{Step 4: Structural Monitoring \& Early-Stopping}
        \STATE $\mathcal{S}_i^{(t)} \gets \text{StopScore}\left( \mathcal{H}(z_{G}^{(t)} \odot M_{i,G}), \mathcal{H}(z_{i,E}^{(t)} \odot M_{i,E}) \right)$
 
        \IF{$\mathcal{S}_i^{(t)} \ge \sigma$ \OR $T-t \ge S$}
            \STATE $\mathcal{A} \gets \mathcal{A} \setminus \{i\}$ \COMMENT{Halt injection after alignment or $S$ steps}
        \ENDIF
    \ENDFOR
    \STATE // \textbf{Step 5: Local Feature Injecting}
    \STATE $\epsilon_{G}^{(t)} \gets \text{DiT}_{G}(z_{G}^{(t)}, \{ f_{i, E}^{(t, k)} \}_{k \in \mathcal{K}, i \in \mathcal{A}}, \{M_{i,G}\}_{i\in\mathcal A}, \lambda, P, C_G)$ \COMMENT{Masked fusion by \Cref{eq:ofi}}
    \STATE // \textbf{Final Global Update}
    \STATE $z_{G}^{(t-1)} \gets \text{Scheduler}(z_{G}^{(t)}, \epsilon_{G}^{(t)}, t)$
\ENDFOR
\RETURN Final image $x = \text{Decoder}(z_{G}^{0})$
\end{algorithmic}
\end{algorithm}

\section{Societal Impacts}
\label{social}
Our proposed framework enables precise, spatially-conditioned image generation with multi-instance control, benefiting applications like creative design and synthetic dataset construction. However, such capabilities could be exploited to produce highly convincing, misleading content or sophisticated deepfakes. Additionally, as our approach builds upon existing generative foundation models, it inevitably inherits the societal biases and stereotypes embedded in their pre-training data. While we focus on addressing challenges in controllable generation, we emphasize that responsible deployment and the development of robust safeguards against privacy, bias, and fairness risks are essential future endeavors.


\end{document}